\documentclass{article}

     \usepackage[final]{neurips_2019}

\usepackage[colorinlistoftodos,prependcaption,textsize=tiny]{todonotes}
\usepackage[utf8]{inputenc} 
\usepackage[T1]{fontenc}    
\usepackage{hyperref}       
\usepackage{url}            
\usepackage{wrapfig,booktabs}
\usepackage{amsfonts}       
\usepackage{nicefrac}       
\usepackage{microtype}      
\usepackage{amsmath}
\usepackage{diagbox}
\usepackage{array,multirow,graphicx}

\newcolumntype{P}[1]{>{\centering\arraybackslash}p{#1}}
\newcolumntype{M}[1]{>{\centering\arraybackslash}m{#1}}
\newcolumntype{?}{!{\vrule width 1pt}}

\title{Blood lactate concentration prediction in critical care patients: handling missing values}

\author{%
    Behrooz Mamandipoor, Mahshid Majd, Monica Moz \thanks{also affiliated with Humanitas Research Hospital, Italy in the capacity of Adult Cardiac Surgeon}  , Venet Osmani \\
  Fondazione Bruno Kessler Research Institute\\
  Trento, Italy\\
  \texttt{\{bmamandipoor, mmajd, mmoz, vosmani\}@fbk.eu} \\
}

\begin{document}

\maketitle

\begin{abstract}
Blood lactate concentration is a strong indicator of mortality risk in critically ill patients. While frequent lactate measurements are necessary to assess patient's health state, the measurement is an invasive procedure that can increase risk of hospital-acquired infections.
For this reason we formally define the problem of lactate prediction as a clinically relevant benchmark problem for machine learning community so as to assist clinical decision making in blood lactate testing. Accordingly, we demonstrate the relevant challenges of the problem and its data in addition to the adopted solutions. Also, we evaluate the performance of different prediction algorithms on a large dataset of ICU patients from the multi-centre eICU database. 
More specifically, we focus on investigating the impact of missing value imputation methods in lactate prediction for each algorithm. The experimental analysis shows promising prediction results that encourages further investigation of this problem.

\end{abstract}


\section{Introduction}
Blood lactate concentration is a biochemical indicator of tissue oxygen delivery and extraction, measured through a blood gas analyser. Inadequate tissue oxygenation results in increased lactate generation. In healthy individuals there is a continuous cycle of lactate production and clearance (metabolised primarily in the liver), while in critically ill patients lactate metabolism is impaired, resulting in elevated lactate concentration. Elevated blood lactate levels are correlated with hospital mortality \citep{nichol2010relative, husain2003serum}, therefore frequent lactate measurements are necessary to track and assess patients' state. However, lactate concentration cannot be measured without drawing arterial or venous blood, which is an invasive procedure that can increase risk of infections. As a consequence, blood gas analysis may not be ordered as frequently leading to sub optimal rates of lactate measurements \citep{rhee2015lactate}. 

In this respect, there is ample potential for machine learning methods to play a significant role in lactate guided clinical decision making. Such role is especially important when considering that lactate-guided therapy significantly reduced hospital mortality and length of stay as evidenced by several multicentre randomised controlled trials \citep{jansen2010early,jones2010lactate,tian2012effect}.
However, this potential has not been explored thus far.

Addressing the evident need for prediction of blood lactate concentration, we define a clinically useful problem \citep{wiens2019no} and compare the performance of several machine learning methods in lactate prediction. Furthermore, we investigate the effect of imputation strategies and missing values on prediction performance. We evaluate the influence of missing value imputation and handling on lactate concentration prediction using 13,464 patients (containing 12,196,798 clinical records) extracted from multi-centre eICU critical care database \citep{pollard2018eicu}.
The main contributions of our work are as follows:
\begin{itemize}
    \item[--] We define a clinically useful problem in critical care, that of blood lactate concentration prediction, and investigate it using machine learning methods;
    \item[--] We compare performance of different algorithms on lactate prediction using a large patient dataset collected from multiple hospitals and ICU units;
    \item[--] We investigate the effect of missing value imputation and handling methods on performance of both traditional machine learning and deep learning algorithms on the defined problem;
    \item[--] Our work can serve as a baseline for future research work to build on top of our results and further advance decision making in critical care.
\end{itemize}

The source code for our experiments will be made public at our GitHub repository so that anyone with access to eICU database can replicate our experiments or build upon our work. 


\section{Methods}\label{proposed}
The first step to successfully deploy machine learning algorithms in healthcare is identifying the right problem to focus on, where the right problem should be of clinical relevance and have appropriate data \citep{wiens2019no}.
Accordingly, we first define the lactate concentration prediction problem and then investigate the data and its challenges. Finally, the well-known prediction models are introduced as baseline methods.
\subsection{Problem definition}
We formally define the problem of lactate concentration prediction as follows: 

For each patient with the set of clinical parameters $S =\{(l,X)_t;t \in T\}$ where $l$ represents blood lactate concentration, $X$ represents the set of all other clinical measures, and $t$ is the time index, we want to predict $l_{t+\beta}$ based on $\hat{S}_t =\{(l,\hat{X})_{t'}; t' \in [{t-\alpha}: t] \}$ where $\hat{X}$ is the selected set of clinically relevant measurements out of $X$. Therefore, we formulate a regression problem using equation \ref{eq:problem}.
\begin{equation}\label{eq:problem}
    \begin{aligned}
   f:\hat{S}_{t} \rightarrow l_{t+\beta} \\
\min_{\theta}Loss(f(\hat{S}_{t},\theta),l_{t+\beta})  
    \end{aligned}
\end{equation}
In other words, the objective of lactate prediction is to predict the blood lactate concentration of a patient in the next $\beta$ hours using a selected set of their measurements taken in the past $\alpha$ hours. 

\subsection{Dataset}

For our work we use eICU Collaborative Research Database \citep{pollard2018eicu}, a multi-center intensive care unit database with high granularity data for over 200,000 admissions to ICUs monitored by eICU programs across the United States. The eICU database comprises 200,859 patient unit encounters for 139,367 unique patients admitted between 2014 and 2015 to hospitals located throughout the US. 
We selected adult patients only with at least 2 lactate measurements that stayed at an ICU unit for more than 18 hours. 

The final patient cohort contained 13,464 patients (14,477 ICU stays) with 12,196,798 clinical records, where we grouped these records into 2 hour windows. Patient's mean age was 61.8 years (45\% female). The complete process of data preparation is described in Section \ref{results}.

\begin{wraptable}[18]{L}[0pt]{0cm}


\begin{tabular}{  c | c  c  c  c }

Age  &   \multicolumn{2}{r}{61.8$\pm$15.7 }\\
LoS* (day)  &  \multicolumn{2}{r}{6.8 $\pm$9.3}\\
\midrule
Male & \multicolumn{2}{r}{7391 (55\%)}\\
Female & \multicolumn{2}{r}{6073 (45\%)}\\
\midrule
Alive* & \multicolumn{2}{r}{10906 (81\%)}\\ 
Dead* & \multicolumn{2}{r}{2558 (19\%)}\\
\midrule
\multicolumn{1}{c}{Top Diagnosis} \\
\midrule
Sepsis, Pulmonary & \multicolumn{2}{r}{1634 (12.1\%)}\\ 
Cardiac arrest & \multicolumn{2}{r}{1155 (8.5\%)}\\ 
Sepsis, Renal/UTI & \multicolumn{2}{r}{859 (6.3\%)}\\ 
Sepsis, GI & \multicolumn{2}{r}{714 (5.3\%)}\\ 


\end{tabular}
\caption{Patient cohort characteristics (*LoS - length of stay, *discharge status)}
\label{tab:eicu}
\end{wraptable}

\subsection{Challenges}

Critical care is an especially data-intensive field, as continuous monitoring of patients in Intensive Care Units (ICU) generates large streams of data. This stream of data creates a great opportunity for the machine learning community and reciprocally machine learning can enhance decision making in critical care \citep{johnson2016machine}. 
However, critical care data suffers from several limitations. 
The main challenges of ICU data can be categorised into three groups: the challenge of data acquisition, the corruption of data, and inherent complexity of ICU data \citep{johnson2016machine}. 
Thanks to the recent availability of large ICU databases such as eICU \citep{pollard2018eicu} and MIMIC III \citep{johnson2016mimic}, the first challenge is becoming less severe. 

\begin{figure}[h!]
\center
\includegraphics[width=0.9\linewidth,scale=0.8]{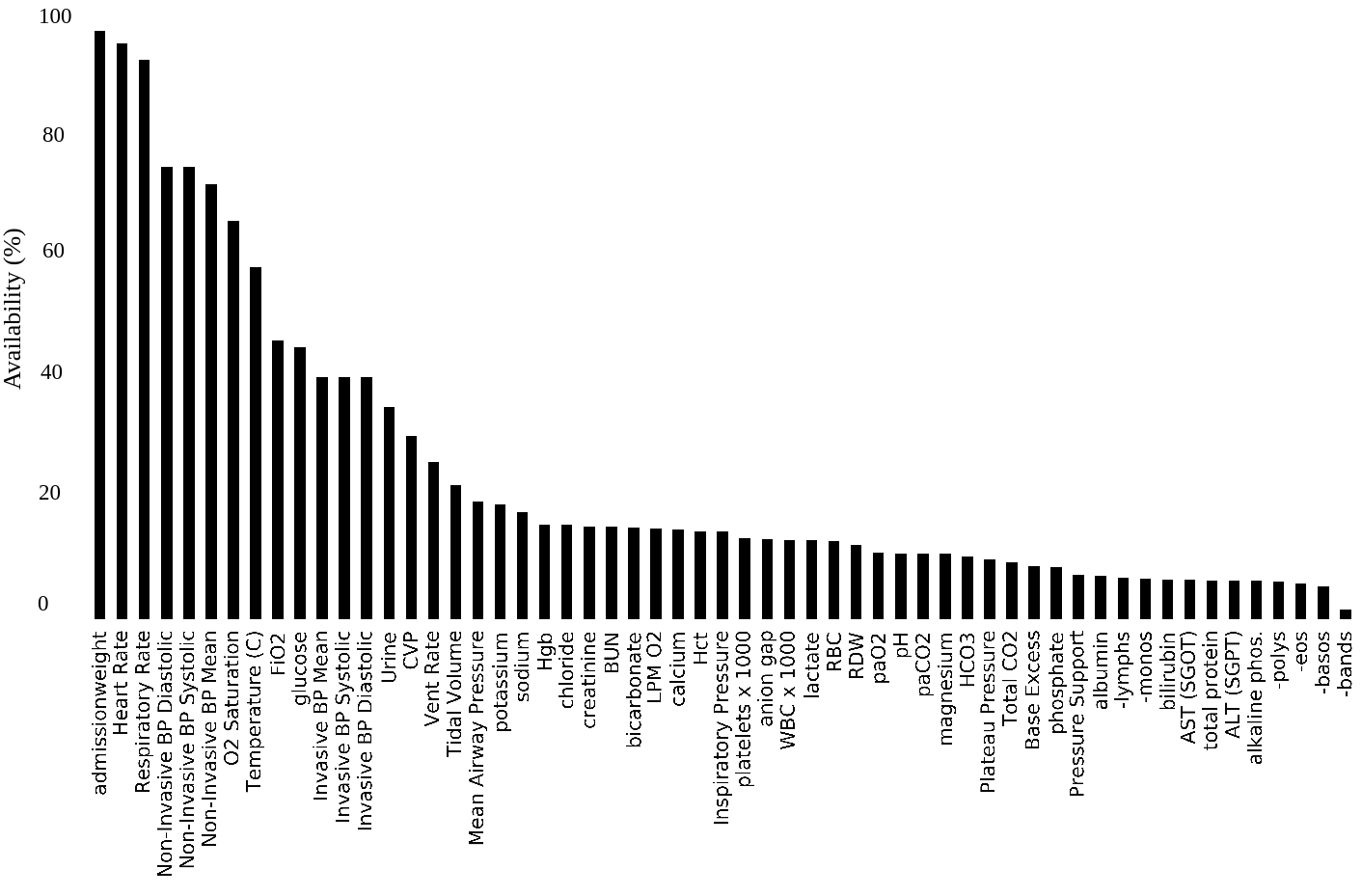}
\caption{The selected features of eICU dataset and the percentage of their observed (available) values. Age, Gender, Ethnicity, and Admission diagnosis are completely available and therefore not included in the plot.}
\label{fig:variables} 

\end{figure}

The corruption of data mainly refers to noise and missing values. It is common to handle noise based on domain knowledge of statistical rules-of-thumb by simply eliminating the unreasonable values. Missing values on the other hand has attracted a lot of attention since most machine learning methods do not function with them \citep{stiglic2017challenges}. Figure \ref{fig:variables} shows the percentage of observed data in the relevant features of eICU dataset regarding lactate prediction. 
The simplest way to handle missing values is to only consider the complete cases, called complete case analysis. Although simple, based on the missing type and rate it could be effective \citep{hughes2019accounting}. In our case the missing rate is so high that complete case analysis is not possible.
Addressing missing values is typically dependent on the cause of missingness.
Statistically speaking, there are three types of missing values: missing completely at random (MCAR) which happens on an unrelated cause, missing at random (MAR) which implies a relation between missing value and other present values, and missing not at random (MNAR) which implies a relationship between the value of the variable and its missingness \citep{rubin1976inference}.

ICU data can contain all types of missingness. Malfunction of machines or human mistakes in ICU can cause MCAR in data. For example, if the material of a test is accidentally broken, the test results would be missing completely at random. This way, missingness is not related to either observed or missing data. MAR is also unavoidable since assessment of some measures are conditioned on the value of other measures. In other words, availability of clinical test may be dependent on patient observation or results of previous tests. In other words, missing at random depends on the observed data.
The last but not the least cause of missingness is difference in frequency of assessing each measure. For example some measures are assessed on an hourly basis, such as blood pressure, and some are performed only daily, such as lab tests. Since this type of missingness is informative about the variable itself, it can be considered as MNAR. Another case for MNAR is that normal values are very likely to be missed in some variables, such as individuals' normal weight.

The third category of challenges reflect the inherent complexity of ICU data which machine learning methods will be responsible to handle. Specifically, we call your attention to two considerable challenges in the introduced lactate prediction problem: complex temporal relations and imbalanced data. The ICU data is longitudinal and each measure can be a time series with varied length and frequency. These time series are not stationary and irregular in time since performing the measurements depends on the status of patient and decision of caregiver.

Predicting blood lactate concentration is a particularly challenging problem as distribution of lactate measurements follows a long-tailed distribution, resulting in a highly imbalanced dataset. Lactate concentration levels for critically ill patients can be divided into four categories: normal (0 to 2.00 $mmol/L$), mild (2.01 to 4.00 $mmol/L$), moderate (4.01 to 6.00 $mmol/L$) and severe hyperlactemia (> 6.01 $mmol/L$) \citep{nichol2010relative}.


\begin{wraptable}[8]{L}[0pt]{0cm}


\begin{tabular}{  l  c  c  c }


0 to 2.00 $mmol/L$  &  46212  &  53.1\% \\

\midrule

2.01 to 4.00 $mmol/L$  &  23037  &  26.5\% \\

\midrule

4.01 to 6.00 $mmol/L$  &  7769  &  8.9\% \\

\midrule

> 6.01 $mmol/L$  &  9979  & 11.5\% \\


\end{tabular}
\caption{Distribution of blood lactate concentration across patients}
\label{tab:lactate_distribution}
\end{wraptable}
As shown in Table \ref{tab:lactate_distribution} almost 80\% of lactate readings are concentrated within normal and mild levels, rendering the problem of predicting moderate and severe hyperlactemia particularly challenging. Furthermore, there is a low linear correlation between lactate levels and relevant clinical parameters (shown in Appendix \ref{appendix}, Figure \ref{fig:correlation}).\\

\subsection{Imputation methods}

The proper imputation method for lactate prediction should consider all forms of missingness in ICU data. Mean, Group mean \citep{sim2015missing}, median, zero and forward imputation are basic single-value imputation methods, which only consider the information in the past values of the variable. Therefore, they work best for data under MCAR assumption and introduce bias and loss of information under MAR or MNAR cases. Yet, these methods are very popular because of their simplicity to implement and interpret. 

Multiple imputation \citep{azur2011multiple}, K Nearest Neighbours \citep{batista2002study}, Matrix Factorisation \citep{koren2009matrix}, PCA \citep{josse2012handling}, SoftImpute \citep{mazumder2010spectral} and Random Forest \citep{stekhoven2011missforest} are the most known traditional machine learning methods that find the substitute for missing values based on the relation between observed and missing features. These methods perform best when handling MAR cases. 
In MCAR case the missingness is completely at random but the missing value might have a relation with other present features; therefore MCAR cases can also benefit from these methods.

Finally, MNAR cases are the hardest to manage. Resampling and alignment are widely used to correct the difference of frequencies in measurements and their uneven alignments. Even after that, a lot of missing values remain. A helpful solution to capture the information behind missingness of variables is to use missing indicators. These indicators can either be used directly aside data or inside the prediction model structure. Note that indicators can also be useful in MAR and MCAR, since they differentiate the imputed values from the observed ones and the prediction method can use this information to ignore bad imputations.


\subsection{Prediction models}

There are a vast number of successful regression methods used for healthcare data which could be grouped into three categories: the statistical regression methods, the traditional machine learning methods and the rising deep learning methods.
Since this problem is basically a regression problem of a real value, it is only logical to apply statistical regression methods as a baseline. The statistical regression methods investigate the relationship between the target variable and other variables. There are several types of statistical regression methods including linear, polynomial, step-wise, ridge, and lasso. Here we apply lasso regression to benefit from its abilities in feature selection and reducing over-fitting. Yet, this method is linear and can only model linear relations in data.

To model the nonlinear relationships in the data, machine learning offers a wide range of methods including support vector regression, back propagation neural networks, K nearest neighbours and decision trees. Among these, random forest (RF) which is an ensemble of decision trees, has shown great performance with clinical data. This may be because RF is able to learn complex and highly non-linear relationships in data and yet it is easily interpretable.

Another informative characteristic of the ICU data is its temporal relations which is not naturally covered by either lasso regression or random forest. 
Recurrent Neural Networks (RNN) address the temporal relation of data by maintaining an internal state. LSTM and GRU are the well-known RNN models which use gating mechanism to avoid gradient explosion/vanishing. Consequently, they are the preferred choice for ICU data.

\section{Results and discussion}\label{results}
In this section the introduced lactate prediction problem is investigated using several imputation methods and prediction models. The parameter $\alpha$ of the task (introduced in Section \ref{proposed}), is set to 6 hours based on the recommendation of SSC on serial lactate testing \citep{levy2018surviving} and $\beta$ is set to 2. In other words, we want to predict the next record (2 hours) of lactate level for a patient based on their available data from at least 6 previous hours.
To do so, all details of the data preparation and parameters of both imputation methods and prediction methods are provided. 

%

%
\textbf{Data preparation.} Data is prepossessed in six steps. (1) Patient cohort is selected based on three inclusion criteria; adult patients (age > 18), with at least two measured lactate levels, and with at least 18 hours length of stay in ICU. (2) The relevant variables are selected based on advice from the clinician. The eICU database consists of 31 different tables which document multiple aspects of each patient’s care such as doctor notes, laboratory tests, active problems, treatments planned, and more \citep{pollard2018eicu}. For our case study, the relevant clinical features for the selected cohort are outlined in Table \ref{table:features}. A detailed description of each variable is provided in the original eICU paper \citep{pollard2018eicu}. The Pearson correlation of the lactate and selected variables is provided in Appendix \ref{appendix}. (3) Some features exist in more than one table under different names. These features are aligned to a unique feature in time.
(4) The selected data is aligned in time since each feature is measured in an arbitrary time and frequency. We resampled time-series data into regularly aligned periods where each feature is sampled every 2 hours. In case a feature is measured more than once during each two hour interval, the last record is used.
(5) Noise and outliers are addressed as follows: for each feature, the valid interval is defined based on clinical knowledge and the values out of the valid scope are considered as missing values.
(6) The data is split into training and testing parts, which is done using five-fold cross validation where at each fold 80\% of the data is considered as training and the rest is test data (the diagram in Appendix \ref{appendix}, Figure \ref{fig:pipeline_diagram} outlines cohort selection, data preprocessing and analysis pipeline).



\begin{table}[!h]
\centering
\caption{List of selected variables from eICU dataset based on clinical relevance (several variable types are recorded in multiple tables)}
\label{table:features}

\resizebox{13cm}{!} 
{

\begin{tabular}{ | >{\centering\arraybackslash}M{1in}  p{10.5cm}|}
\toprule
\hfil \textbf{eICU Tables} & \hfil \textbf{Variables}\\
\bottomrule

Patient &  gender, age, ethnicity, admissionweight, apacheadmissiondx \\
\midrule

Lab & Respiratory Rate, O2 Saturation, FiO2, glucose, potassium, sodium, Hgb, chloride, creatinine, BUN, bicarbonate, LPM O2, calcium, Hct, platelets x 1000, anion gap, WBC x 1000, lactate, RBC, RDW, paO2, pH, paCO2, magnesium, HCO3, Total CO2, Base Excess, phosphate, Pressure Support, albumin, -lymphs, -polys, -eos, -basos, -bands, -monos, bilirubin, AST (SGOT), ALT (SGPT), total protein, alkaline phos.\\
\midrule

NurseCharting &  Heart Rate, Respiratory Rate, Temperature (C), Invasive BP Mean, Invasive BP Systolic, Invasive BP Diastolic, Non-Invasive BP Mean, Non-Invasive BP Systolic, Non-Invasive BP Diastolic, O2 Saturation, glucose, CVP, LPM O2, Total CO2\\
\midrule

RespiratoryCharting &  Heart Rate, Respiratory Rate, FiO2, Total CO2, Tidal Volume, Inspiratory Pressure, LPM O2, Vent Rate, Plateau Pressure, Mean Airway Pressure, Pressure Support, Inspiratory Pressure\\
\midrule

VitalPeriodic &  Heart Rate, Respiratory Rate, Temperature (C), Invasive BP Mean, Invasive BP Systolic, Invasive BP Diastolic, CVP\\
\midrule

VitalAperiodic & Non-Invasive BP Mean, Non-Invasive BP Systolic, Non-Invasive BP Diastolic \\
\midrule
IntakeOutput &  Urine  \\
\bottomrule
\end{tabular}
}
\end{table}

\textbf{Imputation settings.} To investigate the impact of different imputation methods on prediction, we selected a wide range of methods based on their popularity in previous research. List of the selected methods and their description is provided in Appendix \ref{appendix},  Table \ref{table:imputation}. Mean, Median and FeedForward are selected from a single imputation group. Also Group mean, based on severity scale of previous lactate (as shown in Table \ref{tab:lactate_distribution}), is investigated. PCA, Matrix Factorization (MF), SoftImpute, KNN, MICE, and MissForest are picked from traditional machine learning solutions and missing indicators, while Auto-Encoder (AE) are examples of recent imputation methods. Generally, there are two types of architectures available for autoencoders \citep{charte2018practical}: overcomplete, where there are more nodes in hidden layers rather than input layer and undercomplete, where hidden layers have fewer nodes than input layer. Our architecture for AE is similar to \citep{gondara2017multiple}. As autoencoder needs a complete dataset for initialisation, missing values in training and test data are pre-imputed by mean and zero respectively. All the methods were applied using python based on fancyimpute, predictive\_imputer, and sklearn open-source libraries. Also for all the machine learning methods, the default parameters proposed by their authors are used.

\textbf{Prediction settings.} Among the introduced prediction models, we compared the results of Lasso regression (LR), Random Forest (RF) and LSTM as is common practice. Since each patient has different number of ICU records, the data samples have different lengths as well. Lasso regression and Random Forest can not handle different sizes of samples and therefore all samples are zero padded to have the same size. To ease LSTM convergence, data is normalised to have zero mean and a standard deviation of one. The LSTM network has 2 layers of 1024 units with \textit{Glorot} normalisation and \textit{tanh} activation, each followed by a drop out layer of 0.6. Adam optimizer is used with learning rate starting from 0.0001 and the model is trained for 20 epochs with batch size 100. The LSTM model is implemented using Keras and Tensorflow as backend and one GTX 2080 Ti as GPU.

\textbf{Evaluation Metrics.} To measure the quality of imputation methods regarding lactate prediction while preserving the structure of data (no artificial missing values are added), we report Mean Absolute Error (MAE), Root Mean Squared Error (RMSE), and R-squared ($R^2$) as indicators of the predictive performance of the regression models. MAE measures the average magnitude of the errors in a set of predictions and RMSE is a quadratic scoring rule that measures the average square of the error. RMSE has the benefit of penalising large errors more and MAE is more interpretable. Both metrics are negatively-oriented scores. On the other hand R-squared measures the percent of variance explained by the model. This measure calculates how good is the model compared to naive mean model and it is positive oriented. We also report the standard deviation over cross validation folds.

\begin{figure}[h!]
\centering
\includegraphics[width=0.8\linewidth,scale=0.8]{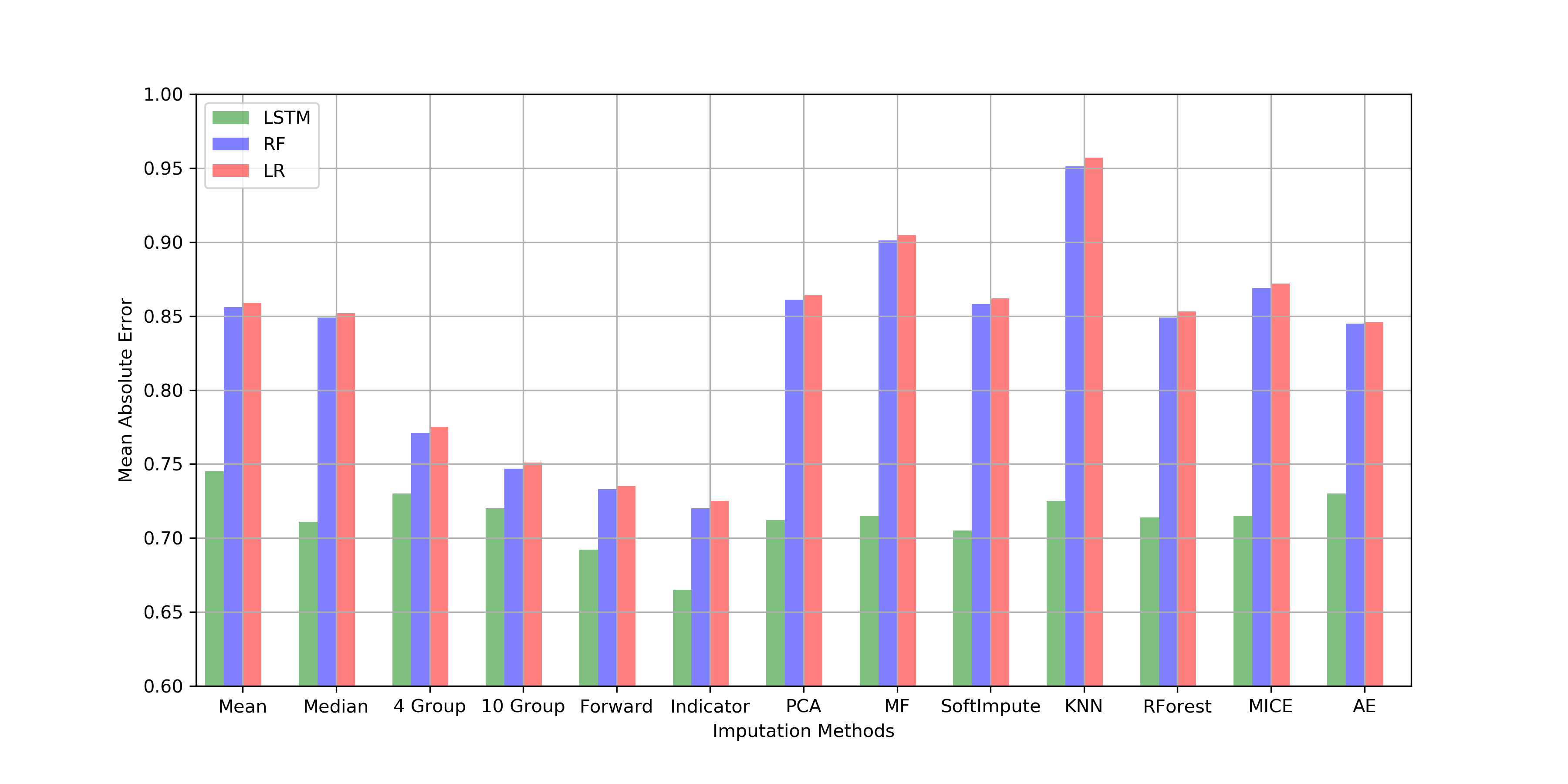}
\caption{The visual plot of the results provided in Table \ref{table:results}. The models are presented in colors and the imputation methods are shown on the horizontal axis.}
\label{fig:compare} 
\end{figure}

\textbf{Results.} All combinations of introduced prediction models and imputation methods are examined on eICU data and their results are reported in Table \ref{table:results}. The mean and standard deviation of each measure on five fold cross-validation of data is reported. The best results are shown in bold. To have a better understanding of the results, the MAE is plotted for all models and methods in Figure \ref{fig:compare}. The models are presented in colours and the imputation methods are in horizontal axis. As the results suggest, both regression model and imputation method affect the prediction results. LSTM performed significantly better compared to LR an RF using the same imputation method. Therefore, it can be concluded that the data contains complex relations not only between measurements but also across their values in time. On the other hand, missing values are best handled using Indicators and Mean method which indicates that missingness is meaningful and happens under MNAR as well as MAR and MCAR assumptions. Forward imputation also shows good results which is related to the fact that different frequency of measurement is the reason behind most of the missingness in ICU data. Since the frequency of a measurement is based on the importance of change in the value of that variable, it makes sense to use the last measured value between the measurement periods. 

\begin{table}[h!]
\caption{The results of all combinations of prediction models and imputation methods. The mean and standard deviation of various metrics on five fold cross validation of data is provided. MAE, RMSE, and $R^2$ stand for Mean Absolute Error, Root Mean Squared Error, and R-squared respectively. LR, RF, and LSTM stand for Linear Regression, Random Forest, and Long Short Term Memory. The imputation methods are described in Appendix \ref{appendix}, Table \ref{table:imputation}.}
 \label{table:results}
\centering
\resizebox{11cm}{!} 
{
\begin{tabular}{c |c| c c c}

\toprule



Measure & \rule{0pt}{2ex} \backslashbox{Imputation}{Regression} & \multicolumn{1}{|c|}{LR} &  \multicolumn{1}{c|}{RF} &  LSTM \\

\midrule

\parbox[t]{2mm}{\multirow{12}{*}{\rotatebox[origin=c]{90}{\textbf{MAE}}}}
& Mean       &  0.859 $\pm$ 0.006 &   0.856 $\pm$ 0.007 &   0.745 $\pm$ 0.045\\
& Median     &  0.852 $\pm$ 0.007 &   0.849 $\pm$ 0.009 &   0.711 $\pm$ 0.006\\
& Groups mean(4)   &  0.775 $\pm$ 0.006 &   0.771 $\pm$ 0.007 &   0.730 $\pm$ 0.004\\
& Feed Forward& 0.735 $\pm$ 0.008 &   0.733 $\pm$ 0.009 &   0.692 $\pm$ 0.008\\
& Indicator  &  0.725 $\pm$ 0.009 &   0.720 $\pm$ 0.010 &   \textbf{0.665 $\pm$ 0.009}\\
& PCA        &  0.864 $\pm$ 0.006 &   0.861 $\pm$ 0.008 &   0.712 $\pm$ 0.008\\
& MF         &  0.905 $\pm$ 0.009 &   0.901 $\pm$ 0.010 &   0.715 $\pm$ 0.012\\
& SoftImpute &  0.862 $\pm$ 0.010 &   0.858 $\pm$ 0.011 &   0.705 $\pm$ 0.006\\
& KNN        &  0.957 $\pm$ 0.010 &   0.951 $\pm$ 0.010 &   0.725 $\pm$ 0.006\\
& MissForest &  0.853 $\pm$ 0.008 &   0.849 $\pm$ 0.005 &   0.714 $\pm$ 0.010\\
& MICE       &  0.872 $\pm$ 0.007 &   0.869 $\pm$ 0.008 &   0.715 $\pm$ 0.009\\
& AE         &  0.846 $\pm$ 0.007 &   0.845 $\pm$ 0.008 &   0.730 $\pm$ 0.051\\

\midrule

\parbox[t]{2mm}{\multirow{12}{*}{\rotatebox[origin=c]{90}{\textbf{RMSE}}}}
& Mean       &   1.263 $\pm$ 0.013 &   1.257 $\pm$ 0.015 &   1.120 $\pm$ 0.016\\
& Median     &   1.259 $\pm$ 0.014 &   1.253 $\pm$ 0.016 &   1.100 $\pm$ 0.014\\
& Groups mean(4)   &   1.155 $\pm$ 0.009 &   1.149 $\pm$ 0.011 &   1.113 $\pm$ 0.017\\
& Feed Forward & 1.120 $\pm$ 0.014 &   1.115 $\pm$ 0.015 &   1.075 $\pm$ 0.016\\
& Indicator  &   1.090 $\pm$ 0.013 &   1.085 $\pm$ 0.016 &   \textbf{1.016 $\pm$ 0.025}\\
& PCA        &   1.268 $\pm$ 0.014 &   1.262 $\pm$ 0.016 &   1.104 $\pm$ 0.014\\
& MF         &   1.304 $\pm$ 0.015 &   1.298 $\pm$ 0.016 &   1.100 $\pm$ 0.018\\
& SoftImpute &   1.268 $\pm$ 0.016 &   1.264 $\pm$ 0.017 &   1.095 $\pm$ 0.017\\
& KNN        &   1.357 $\pm$ 0.016 &   1.349 $\pm$ 0.017 &   1.110 $\pm$ 0.011\\
& MissForest &   1.254 $\pm$ 0.016 &   1.248 $\pm$ 0.011 &   1.100 $\pm$ 0.011\\
& MICE       &   1.270 $\pm$ 0.014 &   1.266 $\pm$ 0.015 &   1.105 $\pm$ 0.016\\
& AE         &   1.245 $\pm$ 0.013 &   1.241 $\pm$ 0.015 &   1.103 $\pm$ 0.022\\

\midrule

\parbox[t]{2mm}{\multirow{12}{*}{\rotatebox[origin=c]{90}{\boldmath{$R^2$}}}}
& Mean       &   0.475 $\pm$ 0.010 &   0.480 $\pm$ 0.013 &   0.585 $\pm$ 0.022\\
& Median     &   0.478 $\pm$ 0.009 &   0.483 $\pm$ 0.012 &   0.601 $\pm$ 0.006\\
& Groups mean(4)   &   0.561 $\pm$ 0.010 &   0.566 $\pm$ 0.013 &   0.592 $\pm$ 0.011\\
& Feed Forward & 0.587 $\pm$ 0.008 &   0.591 $\pm$ 0.011 &   0.620 $\pm$ 0.013\\
& Indicator  &   0.605 $\pm$ 0.012 &   0.610 $\pm$ 0.013 &   \textbf{0.660 $\pm$ 0.017}\\
& PCA        &   0.471 $\pm$ 0.011 &   0.476 $\pm$ 0.014 &   0.598 $\pm$ 0.011\\
& MF         &   0.440 $\pm$ 0.013 &   0.445 $\pm$ 0.015 &   0.601 $\pm$ 0.014\\
& SoftImpute &   0.471 $\pm$ 0.010 &   0.475 $\pm$ 0.013 &   0.605 $\pm$ 0.006\\
& KNN        &   0.391 $\pm$ 0.011 &   0.397 $\pm$ 0.016 &   0.595 $\pm$ 0.010\\
& MissForest &   0.482 $\pm$ 0.007 &   0.487 $\pm$ 0.006 &   0.601 $\pm$ 0.012\\
& MICE       &   0.469 $\pm$ 0.010 &   0.472 $\pm$ 0.012 &   0.598 $\pm$ 0.011\\
& AE         &   0.489 $\pm$ 0.010 &   0.493 $\pm$ 0.012 &   0.599 $\pm$ 0.024\\
\bottomrule

\end{tabular}
}
\end{table}

\section{Related Work}
There are no related works concerning lactate prediction; therefore, an overview on the problems defined on EHR data is provided, and then the literature of ICU data challenges are reviewed.
While there is an extensive body of research on machine learning for secondary use of EHR data, the common benchmark problems are limited to mortality detection, length of stay (LOS) prediction, phenotyping and sepsis detection \citep{purushotham2017benchmark, harutyunyan2017multitask}. These problems are well-defined with proper data and baseline results and consequently have been the primary focus of machine learning community for the last decade. 
Beside these, only a small number of other topics with medical relevance are considered by machine learning researchers \citep{zheng2017machine, kawaler2012learning}. Instead, the target has been to improve the results of the well-known benchmarks using different machine learning methods, from traditional to state of the art \citep{shickel2017deep}. 
In \citep{harutyunyan2017multitask}, linear regression and LSTM-based methods are studied on the benchmark problems where unsurprisingly LSTM outperforms the linear models. Purushotham et al. provid the results of a range of methods including linear regression, tree-based models, neural networks and deep learning methods on the famous benchmark problems \citep{purushotham2017benchmark}. Johnson et al. analyze the results of several studies on mortality detection and compare them with linear regression and random forrest  \citep{johnson2017reproducibility}. 

Another common research direction is to understand the challenges of health data and the best ways to handle them \citep{schmitt2015comparison,le2018comparison}. 
Vesin et al. are the first to properly address missing values in ICU data  \citep{vesin2013reporting} reporting that only 4 percent of published manuscripts mention and handle the missing values. 
Cismondi et al. demonstrate the effect of imputation on a real ICU database with 16 features and up to 69\% of missingness \citep{cismondi2013missing}. They used resampling and ignored those missing data that could be explained by other variables. Johnson et al. provided a comprehensive research on ICU data, its challenges and the solutions offered by machine learning methods\citep{johnson2016machine}. Other valuable researches on handling missing values in ICU data includes \citep{lipton2016directly, sharafoddini2019new} where missing indicators are investigated for phenotyping and mortality detection. They concluded that enriching the data with missing indicators lead to prediction improvement. Another direction of study tries to modify the LSTM or GRU units to internally consider indicators and their time intervals \citep{che2018recurrent}. 
 


\section{Conclusion}\label{conclusion}
In this study we have defined a clinical problem that has not been addressed previously, describing the necessity of lactate concentration prediction in critical care decision making. Performance comparison between a number of algorithms shows that LSTM-based method can predict lactate level with a Mean Absolute Error of 0.665 across 13,464 patients from different hospitals and ICU units. Furthermore, we show that indicator imputation method achieves highest performance in our dataset, suggesting that a missing value indicator is informational and increases predictive power over other, mean-based imputation methods. 
Our work is a promising first step towards applying machine learning techniques in predicting lactate concentration while there is apparent room for improvement. 
The presented results can serve as a basis for future work and further investigation of clinical decision making in critical care.
%

\section{Future directions}\label{future_directions}
We are investigating a number of future directions for this research, including: i) validation of our results using external datasets: eICU data is collected from multiple ICU centres across the United States that may result in lower systematic bias in comparison to a single centre dataset, such as MIMIC III; however the latter has a richer collection of pharmacological interventions which may further improve the prediction results, especially for patients with high lactate levels, where pharmacological interventions are more frequent; ii) lactate distribution by nature is highly imbalanced, yet addressing imbalance in a regression task is far less mature than in a classification task \citep{krawczyk2016learning}, where ensemble learning may play a role; iii) categorise lactate levels into four groups according to severity of hyperlactatemia (Table \ref{tab:lactate_distribution}), effectively converting the task into a classification problem and addressing the previous challenge; iv) a promising avenue is to tailor state of the art deep learning methods (such as GRU-D) to capture informative missingness or learn compressed representations of multivariate time series using generative models; and v) model interpretability will play an important role in increasing trust as translating machine learning models effectively to clinical practice requires establishing clinicians' trust.

\bibliographystyle{apalike}
\bibliography{bib_lactate}

\appendix
\section{Appendix}
\label{appendix}
\begin{table}[h!] 
\caption{Selected imputation methods and their description.}
\label{table:imputation}
\centering

\begin{tabular}[ht]{  >{\centering\arraybackslash}M{1in}  p{10.5cm}}
\toprule
\hfil \textbf{Methods} & \hfil \textbf{Description}\\
\bottomrule
Mean and Median & Impute missing values with overall mean or median of observed data for each variable.\\
\midrule
Group mean & Group data based on other observed variables and replace the missing values with the mean of the corresponding group.\\ 
\midrule
Feed Forward &  Propagate measurements forward (or backward) in time to fill missing data using the nearest valid value.\\ 
\midrule
Indicator \& Mean &  Impute with mean of observed data while keeping a binary indicator column for each variable representing if the value is imputed or not.\\
\midrule
PCA &  Fill in missing values using probabilistic principal components analysis (PCA). \\
\midrule
MF &  (Matrix Factorization) Use direct factorization of the incomplete matrix into two low-rank matrices. \\
\midrule
SoftImpute &  Iterative method for matrix completion with iterative soft-thresholded Singular Value Decomposition (SVD) to impute the missing values. \\
\midrule
KNN &  Impute missing values based on the K most similar observed neighbours of missing data. \\
\midrule
MissForest &  Iteratively use Random Forest to create multiple imputations of each variable. \\
\midrule
MICE &  (Multiple Imputation by Chained Equations) Use sequential regression to create several imputations for each feature. \\
\midrule
AE &  (Autoencoder) Reconstruct input data by learning a function using neural network.\\
\bottomrule
\end{tabular}
\end{table}%


\begin{figure}[h!]
\centering
\includegraphics[width=1\linewidth,scale=1]{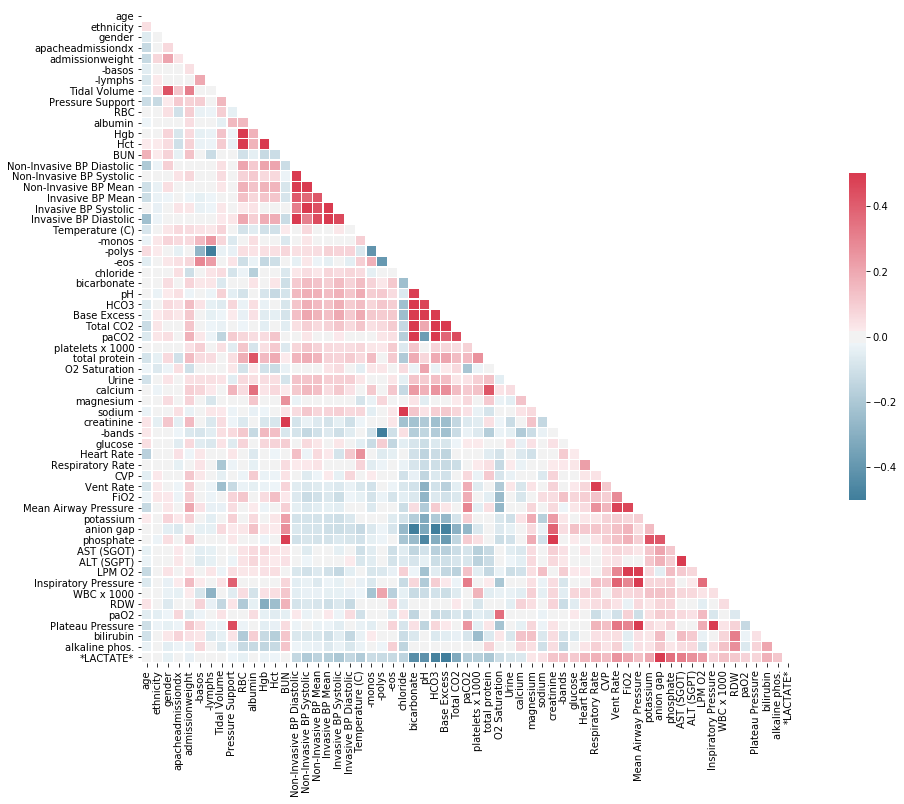}
\caption{Pearson correlation between all the selected variables}
\label{fig:correlation} 
\end{figure}


\begin{figure}[h]
\centering
\includegraphics[width=0.9\linewidth,scale=0.9]{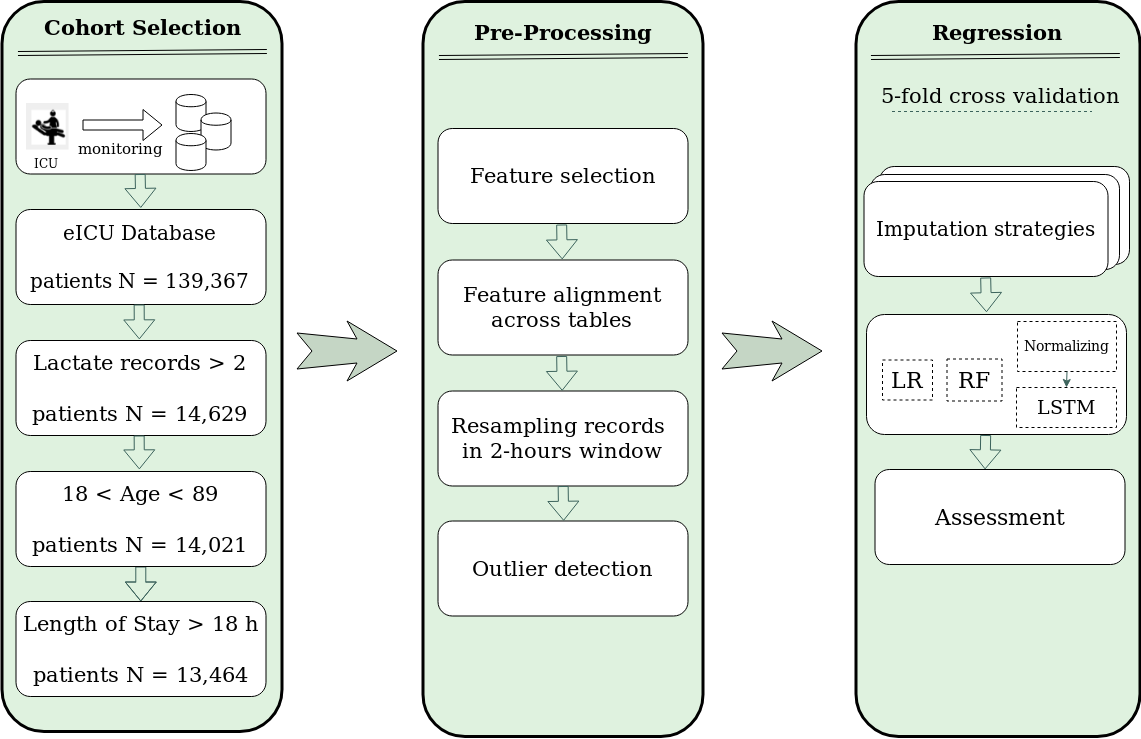}
\caption{Study cohort selection, dataset preparation and model establishment pipeline}
\label{fig:pipeline_diagram} 
\end{figure}

\end{document}